\pdfoutput=1
\documentclass[11pt]{article}

\usepackage[preprint]{acl}

\usepackage{times}
\usepackage{latexsym}

\usepackage[T1]{fontenc}
\usepackage[utf8]{inputenc}

\usepackage{microtype}

\usepackage{inconsolata}

\usepackage{graphicx}
\usepackage{lipsum,multicol}
\usepackage{fancyvrb}
\usepackage{array}
\usepackage{geometry}
\usepackage{enumitem}
\usepackage{tabularx}
\usepackage{booktabs}
\usepackage{ragged2e} % For \RaggedRight in X columns (better than standard \raggedright)
\usepackage{subcaption}  
\usepackage{latexsym}
\usepackage{amsmath, amssymb, amsfonts} 
\usepackage[ruled]{algorithm2e} % For math symbols
\usepackage{algorithmicx}
\usepackage{algpseudocode}
\usepackage{lipsum}
\usepackage{multicol}
\usepackage{booktabs}
\usepackage[most]{tcolorbox}
\tcbuselibrary{breakable,skins}
\usepackage{cuted}

\newcount\Comments  % 0 suppresses notes to selves in text
\definecolor{purple}{rgb}{0.5,0,1}
\definecolor{teal}{rgb}{0.33,0.65,0.55}
\definecolor{green}{rgb}{0.1,0.65,0.1}
\newcommand{\kibitz}[2]{\ifnum\Comments=1\textcolor{#1}{#2}\fi}

\title{TTT-Bench: A Benchmark for Evaluating Reasoning Ability \\ with Simple and Novel Tic-Tac-Toe-style Games}
\author{Prakamya Mishra\thanks{Corresponding author: Prakamya Mishra \\ (prakamya.mishra@amd.com)}, 
   Jiang Liu,
  Jialian Wu,
  Xiaodong Yu, \\
  \textbf{Zicheng Liu},
  \textbf{Emad Barsoum} 
  \\
  Advanced Micro Devices, Inc. (AMD) \\
  \href{https://prakamya-mishra.github.io/TTTBench}{https://prakamya-mishra.github.io/TTTBench}}
\begin{document}
\maketitle
\begin{abstract}
 Large reasoning models (LRMs) have demonstrated impressive reasoning capabilities across a broad range of tasks including Olympiad-level mathematical problems, indicating evidence of their complex reasoning abilities. While many reasoning benchmarks focus on the STEM domain, the ability of LRMs to reason correctly in broader task domains remains underexplored. In this work, we introduce \textbf{TTT-Bench}, a new benchmark that is designed to evaluate basic strategic, spatial, and logical reasoning abilities in LRMs through a suite of four two-player Tic-Tac-Toe-style games that humans can effortlessly solve from a young age. We propose a simple yet scalable programmatic approach for generating verifiable two-player game problems for TTT-Bench. Although these games are trivial for humans, they require reasoning about the intentions of the opponent, as well as the game board's spatial configurations, to ensure a win. We evaluate a diverse set of state-of-the-art LRMs, and \textbf{discover that the models that excel at hard math problems frequently fail at these simple reasoning games}. Further testing reveals that our evaluated reasoning models score on average $\downarrow$ 41\% \& $\downarrow$ 5\% lower on TTT-Bench compared to MATH 500 \& AIME 2024 respectively, with larger models achieving higher performance using shorter reasoning traces, where most of the models struggle on long-term strategic reasoning situations on simple and new TTT-Bench tasks.
\end{abstract}

\begin{figure*}[!ht]
\centering
  \includegraphics[width=0.8\textwidth]{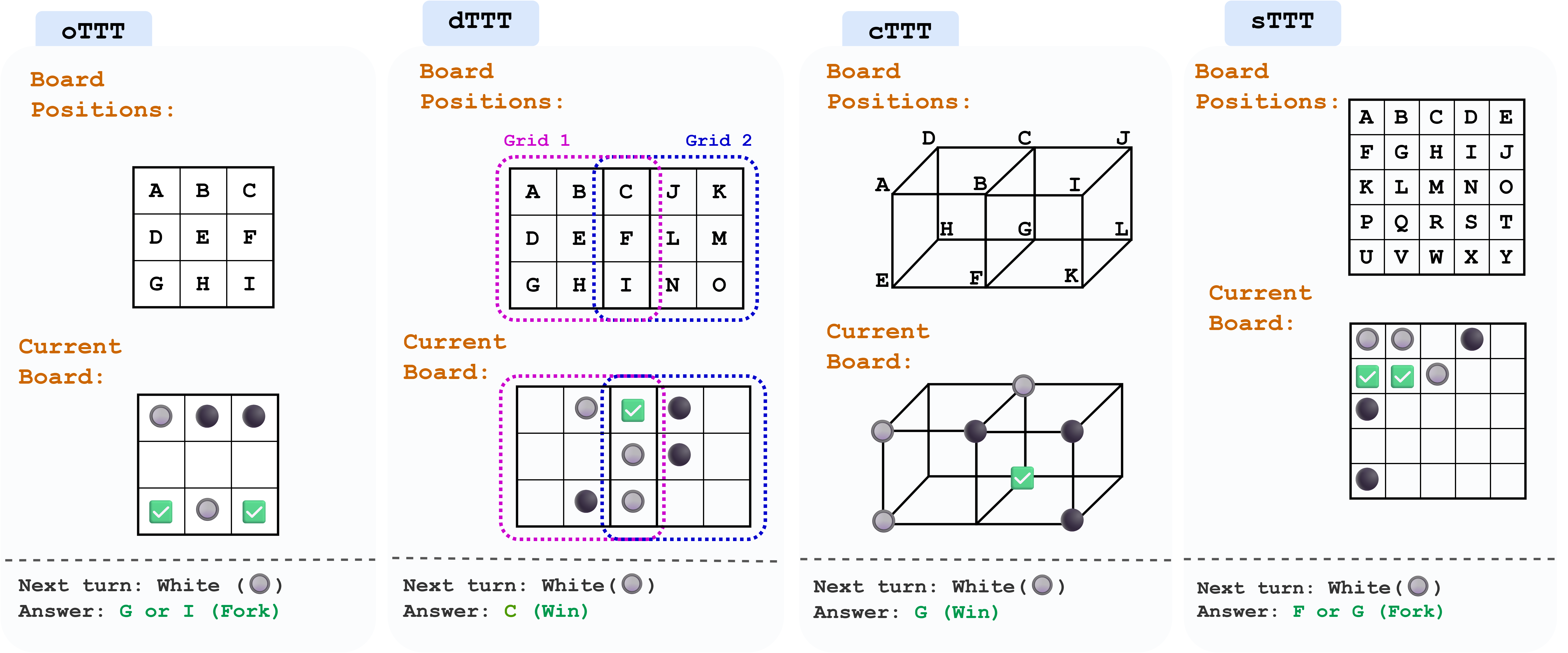}
  \caption{Visualization of four game types in TTT-Bench (oTTT, dTTT, cTTT, and sTTT). For each game, the illustration consists of the board position annotations (above), the current game state with Alice's (white) and Bob's (black), and the correct next move (represented by a check mark).}
  \label{fig:tttbench}
\end{figure*}

\section{Introduction}
Recent advances in large reasoning models (LRMs) have driven significant breakthroughs across various reasoning tasks \citep{jaech2024openai, deepseekai2025deepseekr1incentivizingreasoningcapability, qwq32b, zhao2024marcoo1openreasoningmodels, guan2025rstar}, including deductive, arithmetic, commonsense, relational, and symbolic reasoning \citep{huang-chang-2023-towards}. The high quality of reasoning data and improvements in reinforcement learning techniques \citep{luo2025wizardmathempoweringmathematicalreasoning, NEURIPS2024_76ec4dc3, NEURIPS2023_a85b405e, xu2024magpiealignmentdatasynthesis, grattafiori2024llama3herdmodels} have markedly improved the performance of LRMs on previously challenging mathematics benchmarks, such as GSM8K \citep{cobbe2021trainingverifierssolvemath},  MATH 500 \citep{lightman2024lets}, and AIME 2024 \citep{aime2024}. However, despite these advancements, the ability of LRMs to reason well on a broader set of reasoning problems, in domains other than maths \& STEM, remains largely unexplored. 

Many of the existing reasoning benchmarks focus on mathematical or STEM-related problems \citep{lightman2024lets, he-etal-2024-olympiadbench, rein2024gpqa, suzgun-etal-2023-challenging, zhang-etal-2024-geoeval, mao2024champcompetitionleveldatasetfinegrained}, mainly due to their verifiable nature and the abundance of annotated problems \citep{li202512surveyreasoning, xu2025largereasoningmodelssurvey}. This narrow focus limits the evaluation of LRMs to specific domains and leaves their broader reasoning capabilities largely untested \citep{huang-chang-2023-towards}. There is a clear need for benchmarks that evaluate a diverse set of reasoning capabilities \citep{chang2023surveyevaluationlargelanguage}, including strategic, spatial, and logical reasoning. Exploring this broader range of reasoning capabilities is particularly compelling when we consider human reasoning abilities. Even young children effortlessly perform intuitive reasoning tasks, particularly in strategic two-player games. They naturally anticipate their opponent’s moves, block threats, and plan multi-step strategies, relying on an instinctive grasp of strategy and spatial relationships rather than explicit or formal reasoning.

An open question that emerges is:
\textit{Whether reasoning models proficient in solving sophisticated mathematical problems can correctly reason on broader domains of problems that require basic strategic, spatial, and logical reasoning with straightforward rational and minimal requirements for specialized knowledge.}
Although there are a few efforts to use two-player board games to evaluate these capabilities in language models \citep{fontana2024nicerhumanslargelanguage, qin-etal-2024-uno, costarelli2024gamebenchevaluatingstrategicreasoning, huang2025fardecisionmakingllmsevaluating}, these benchmark games are well-documented in the literature and online, hurting their reliability due to the elevated risk of training of these LRMs on associated benchmark test data. 

Motivated by this gap, we introduce \textbf{TTT-Bench}, a new benchmark specifically created to evaluate the reasoning capability of LRMs through a suite of \textbf{simple and novel two-player Tic-Tac-Toe-style games} (Figure \ref{fig:tttbench}). Although trivial for humans, these games require basic strategic reasoning, including predicting an opponent's intentions and understanding spatial configurations. TTT-Bench is a simple and diverse benchmark consisting of four types of two-player games, namely oTTT, dTTT, cTTT, and sTTT (see Figure \ref{fig:tttbench} for examples), and the questions in the benchmark test the reasoning capability of LRMs by posing the question of predicting the best move by a player. The objectives of these games are not only simple but are also introduced here for the first time (except oTTT), with no documented prior literature available online, making our benchmark uncontaminated and highly reliable. Briefly, for each task in TTT-Bench, two players play a strategic game on a configuration dependent on the task to reach a winning state that satisfies a task-dependent winning constraint, and the player who reaches the winning state first wins the game. The benchmark consists of programmatically generated board game questions along with their verifiable answers, where we employ a simple heuristic-based algorithm (Appendix \ref{aglo:ttt-bench}) to identify the finite set of next best possible moves and their verdict as the answers to ensure the verifiability of the solutions.

The main contributions of this work are as follows:
\begin{itemize}
    \item In this work, we introduce \textbf{TTT-Bench} benchmark to evaluate the reasoning capabilities of LRMs that are proficient in solving difficult math problems, on a broader domain of basic strategic, spatial, and logical reasoning tasks through a suite of four simple two-player Tic-Tac-Toe-style games.
    \item Evaluated a variety of state-of-the-art (SOTA) LRMs on TTT-Bench, and revealed a surprising finding: \textbf{models skilled at difficult math problems frequently struggle with these simpler reasoning tasks}.
    \item This work not only highlights a fundamental shortcoming in existing LRMs, but also provides a new, simple, and scalable approach for automated verifiable two-player game generation, fostering future research efforts in evaluating the reasoning capability of LRMs.
\end{itemize}

\section{Related Work}

\paragraph{Advances in Mathematical Reasoning with Large Language Models}
Recent developments in LRMs have led to significant improvements in their reasoning capabilities \citep{o3-sytem-card,jaech2024openai,deepseekai2025deepseekr1incentivizingreasoningcapability,qwq32b,exaone-deep, team2025kimi} - exhibited by an exponential performance boost in complex reasoning tasks, especially on mathematical problem-solving benchmarks \citep{lightman2024lets,aime2024}. Contrary to massive reasoning models, efficiently trained small reasoning models \citep{deepscaler2025, wen2025light, sky_t1_2025, ye2025limoreasoning, muennighoff2025s1simpletesttimescaling} have also proven to achieve competitive results across these benchmarks. Rapid improvements in reasoning capabilities of LRMs can be attributed to various training methodologies \citep{yu2025dapoopensourcellmreinforcement,shao2024deepseekmathpushinglimitsmathematical}, where Reinforcement learning with verifiable rewards (RLVR) \citep{lambert2024t,deepseekai2025deepseekr1incentivizingreasoningcapability} has emerged as a predominant approach.

\paragraph{Evaluating Reasoning Ability of LRMs}
Despite these advancements, the present widely used reasoning benchmarks are predominantly based on mathematical and STEM-focused tasks, potentially limiting the assessment of these models' reasoning ability in much broader domains. Several works have demonstrated significant shortcomings of LRMs on intuitive reasoning tasks \citep{valmeekam2024planningstrawberryfieldsevaluating, valmeekam2024llmscantplanlrms, electronics13081532,topsakal2024evaluatinglargelanguagemodels}, where even SOTA reasoning models fall short in planning and reasoning. Such findings underscore the necessity for the development of benchmarks that assess a broader spectrum of reasoning, encompassing intuitive and strategic thinking.

\section{TTT-Bench}
TTT-Bench is a simple and scalable two-player Tic-Tac-Toe (TTT) style game benchmark for evaluating the reasoning ability of the present SOTA LRMs. This benchmark consists of four games, namely: \textbf{oTTT} (ordinary TTT), \textbf{dTTT} (double TTT), \textbf{cTTT} (TTT in cube), and \textbf{sTTT} (TTT with squares), which are described in detail below:
\begin{itemize}
    \item \textbf{oTTT:} This is the ordinary TTT game, where two players play a game on a 3x3 grid. The points on the grid are labeled top to bottom and left to right as A, B, C, D, E, F, G, H, and I. One player plays with white stone ($\circ$) and the other uses black stone ($\bullet$). At each turn, the player places a stone of the corresponding color onto one of the positions that has not been occupied. \textbf{\textit{Whoever has three stones in a line (horizontal, vertical, or diagonal) wins.}}
    \item \textbf{dTTT:} This is a variant of TTT with two (double) adjacent 3x3 grids. The points on the first grid are labeled top to bottom, left to right, as A, B, C, D, E, F, G, H, and I. The points on the second grid are labeled top to bottom, left to right, as C, J, K, F, L, M, I, N, and O. Here, note that the points C, F, and I are shared by the two grids. One player plays with white stone ($\circ$) and the other uses black stone ($\bullet$). At each turn, the player places a stone of the corresponding color onto one of the positions that has not been occupied. \textbf{\textit{Whoever has three stones in a line (horizontal, vertical, or diagonal) on either grid wins.}}
    \item \textbf{cTTT:} This is a variant of TTT which is played on two adjacent cubes. The points A, B, C, and D form the top rectangle in the first cube, and B, I, J, and C form the top rectangle in the second cube. Points E, F, G, and H form the bottom rectangle in the first cube, and F, K, L, and G form the bottom rectangle in the second cube. A-E, B-F, C-G, D-H, I-K, and J-L are the edges. Here, note that the points B, C, G, and F are shared by the two cubes. One player plays with a white sticker ($\circ$) and the other uses a black sticker ($\bullet$). At each turn, the player places a stone of the corresponding color onto one of the positions that has not been occupied. \textbf{\textit{Whoever has four stickers on the same plane on either cube wins.}}
    \item \textbf{sTTT:} This is a variant of TTT which is played on a large grid of five intersections between horizontal and vertical lines. There are five equally spaced horizontal lines, where the distance between two neighboring horizontal lines is one. Similarly, there are five equally spaced vertical lines, and the distance between two neighboring vertical lines is one. There are twenty-five intersection points between the five horizontal and vertical lines. These twenty-five points are labeled from top to bottom, left to right, as A, B, C, D, …, Y. One player plays with white stone ($\circ$) and the other uses black stone ($\bullet$). At each turn, the player places a stone of the corresponding color onto one of the twenty-five points that have not been occupied. \textbf{\textit{Whoever has four stones that form either a unit square (with side length of one) or a “diagonal square” with side length equal to the square root of two wins.}}
\end{itemize}

Examples of these games are illustrated in Figure \ref{fig:tttbench}, with the corresponding exact questions listed in the Appendix \ref{Appendix-Examples}. As seen from the illustrations, such games are easy for humans to play even from a young age, where a player's victory is based on their ability to anticipate their opponent’s move, blocking threats, and if they can form basic winning strategies --  skills demonstrating the simple reasoning ability of humans.

All the problems in this benchmark are formulated as a text-based board game question that starts with the description of the game, as well as the current game state with all the moves played by both players. The final TTT-Bench task questions are formulated to predict the next best move position by the current player (Appendix \ref{Appendix-Examples}), like: \texttt{Where should Bob play next?}. Such question formulation ensures that our benchmark solving complexity is low, as just predicting the next move out of a small set of possible available moves in our games guarantees a small solution search space, making our benchmark questions relatively easier compared to other works \citep{illusion-of-thinking}. Since the outcomes of these games are based on finite and deterministic logic, possible next-best moves can be easily identified. Leveraging this, we created an automated pipeline to generate new test samples for each of the four games with verifiable solutions, enabling this benchmark to not only have new and unique questions that the SOTA LRMs are not contaminated against, but also make our benchmark highly scalable in terms of the size, diversity and difficulty.

\subsection{Automated TTT-Bench Generation} \label{sec:auto-gen-bench}
We employ a programmatic approach to generate a diverse set of questions corresponding to the four games present in the TTT-Bench. In this approach, for each game, we first generate all possible game states between two players having played $N$ moves. We then filter out the states for further processing where 1) neither player has already won, and 2) where the player who made the $N^{th}$ move has not yet established a "Winning Fork". Here, by "Winning Fork", we refer to the game state in which a player has two or more potential immediate winning states, with each potential winning state being one turn (move) short of winning the game. In this state, independent of the next move of the opponent, the current player can win with 100\% probability by pursuing either of the winning moves on their next chance. Refer to Appendix \ref{Appendix-Examples} for more illustrative examples of "Winning Fork" states. This leaves us with all the game states in which there is a deterministic set of positions in the board for the player playing the $({N+1})^{th}$ move, that can lead to their victory in the game.

\begin{figure}[ht]
\centering
  \includegraphics[width=0.8\columnwidth]{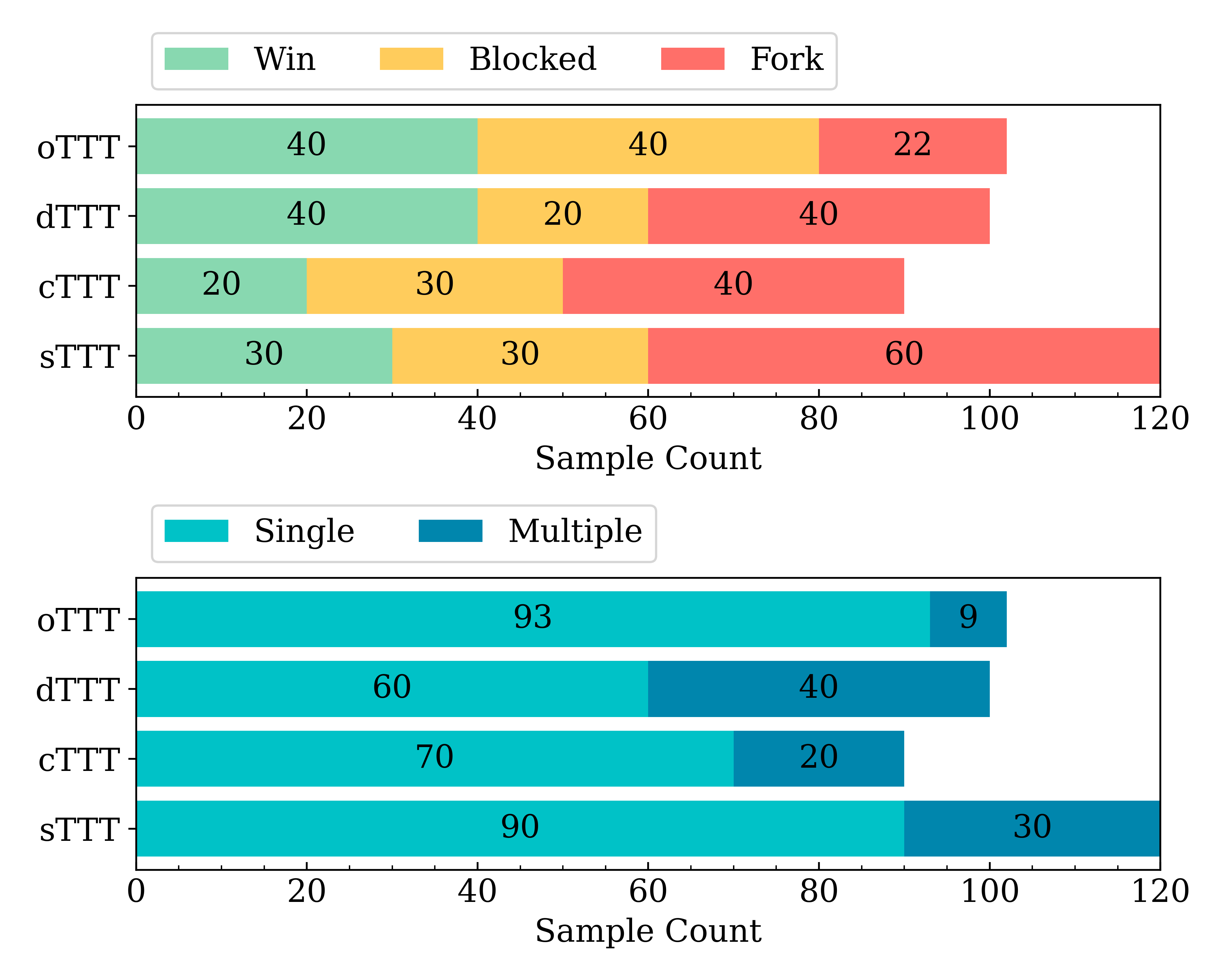}
  \caption{Distribution of TTT-Bench question solution types. \textbf{Above:} Distribution of questions and their game verdicts (Win, Blocked, or Fork) based on how the game concludes after the corresponding solution move is played. \textbf{Below:} Distribution of questions and their solution types (single answer or multiple next-best moves).}
  \label{fig:test-dist}
\end{figure}

\begin{table}[!ht]
\centering
\resizebox{0.8\columnwidth}{!}{%
\begin{tabular}{c|p{8cm}}
\hline
Verdict Category & Description                                                                                                                                \\ \hline
Win              & The player wins the game by playing the $({N+1})^{th}$ move.                                                                                        \\ \cline{2-2}
Blocked          & The player blocks the opponent's immediate potential win by playing the $({N+1})^{th}$ move.                                                        \\ \cline{2-2}
Fork             & The player establishes two or more potential winning states (with exactly one turn short of winning in all the states) by playing the $({N+1})^{th}$ move. \\ \hline
\end{tabular}%
}
\caption{Description of game verdict categories in TTT-Bench games.}
\label{verdict-cats}
\end{table}
\begin{figure*}[!ht]
    \centering
    \includegraphics[width=0.9\textwidth]{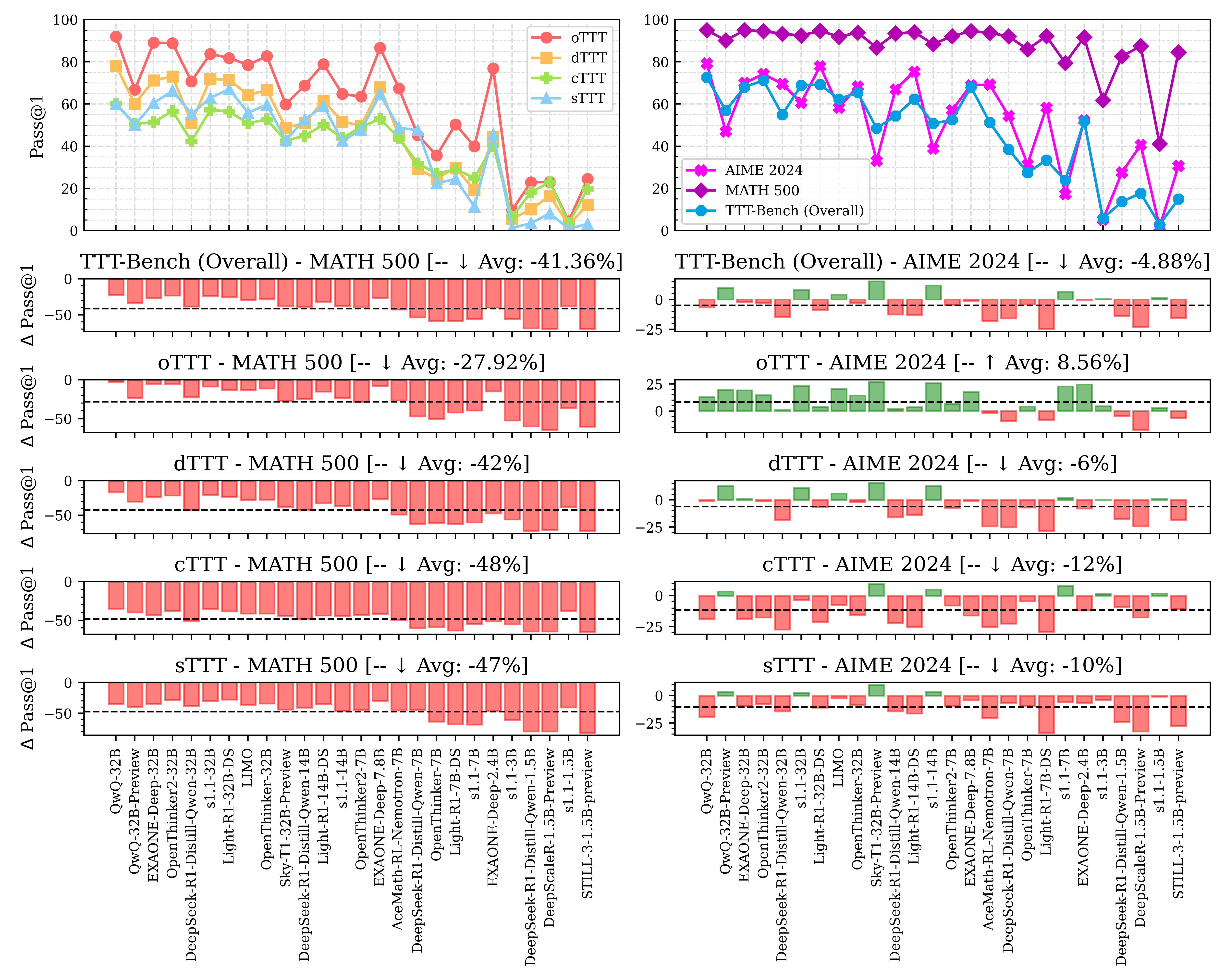}
    \caption{Pass@1 performance comparison of LRMs on TTT-Bench (Left) and math benchmarks: AIME 2024 \& MATH 500 (Right). Exact performance numbers available in Table \ref{main-result-table} of Appendix \ref{appendix-results}}
    \label{fig:test-reasoning-v-math}
\end{figure*}
Our final test set consists of only those games in which $N$ moves had been played, and the verdict of the game after the $(N+1)^{th}$ move falls into one of the categories listed in  Table \ref{verdict-cats}. These categories ensure that the solutions generated for TTT-Bench questions, for all the games, are the optimal next-best moves. The algorithm \ref{aglo:ttt-bench} in Appendix \ref{sec:appendix-algo} describes how we generated optimal next-best move solutions for the TTT-Bench questions, where the game verdicts are either "Win", "Blocked", or "Fork". In the algorithm, we first check if there exists any winning moves at the $(N+1)^{th}$ chance where $N$ moves have already been played so far, and use those moves as the solution with the verdict of "Win". If there are no winning moves, then we checked if there are moves by which the opponent can win on the $(N+1)^{th}$ chance, and used those as the solution with the verdict of "Blocked", since not playing those moves will ensure the opponents win. Finally, in case of no blocking moves as well, we identified moves that result in a fork for the current player that will ensure a future win, and used those moves as the solution with the verdict of "Fork".

Based on this automated game question-answer generation approach explained above, we generate a diverse final test set consisting of \textasciitilde 100 samples from each game. For each game, we used a set of predefined values of $N$ such that the games were simple, solvable, and had a low branching factor. For oTTT we use $N=\{4,5\}$, $N=\{6,7\}$ for dTTT, $N=\{5,6,7\}$ for cTTT, and $N=\{6,7\}$ for sTTT. As shown in Figure \ref{fig:test-dist}, not only is this dataset diverse across different game verdicts, but it also consists of questions with both single and multiple next-best move solutions. Overall, the final TTT-Bench benchmark consists of 412 questions from four simple two-player TTT games. We release this benchmark for future research under the Apache 2.0 License.

\section{Main Results}
\subsection{Evaluation Setup}
\begin{figure*}[!ht]
    \centering
    \includegraphics[width=\textwidth]{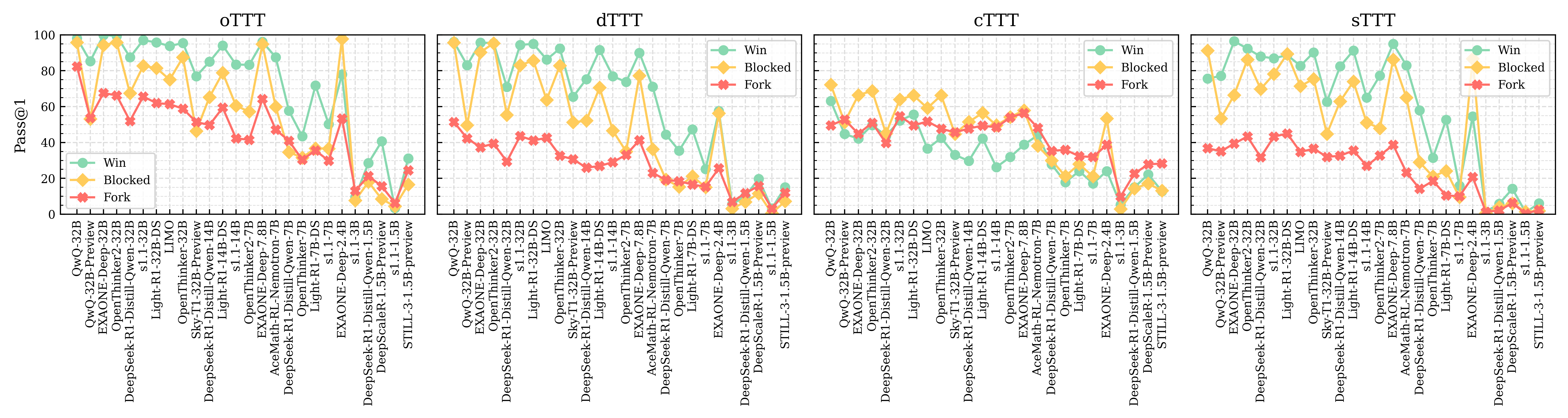}
    \caption{Pass@1 performance analysis of LRMs over individual solution verdict category ("Win", "Blocked", \& "Fork") questions in TTT-Bench tasks. Exact performance numbers available in Table \ref{catergory-results} of Appendix \ref{appendix-results}.}
    \label{fig:test-category}
\end{figure*}

In this work, we evaluate a comprehensive set of recent SOTA LRMs on TTT-Bench and conduct a head-to-head comparison of their performance on TTT-Bench versus two widely used mathematics benchmarks: AIME 2024 \citep{aime2024} (Olympiad-level math) \& MATH500 \citep{lightman2024lets} (high school math), to thoroughly investigate the reasoning capabilities of these models. The parameter count of the models in our evaluations ranges from 1.5 billion to 32 billion parameters, which includes models like DeepSeek-R1-Distill-Qwen-1.5B / 7B / 14B / 32B \citep{deepseekai2025deepseekr1incentivizingreasoningcapability}, S1.1-1.5B / 3B / 7B / 14B / 32B \citep{muennighoff2025s1simpletesttimescaling}, EXAONE-Deep-2.4B / 7.8B / 32B \citep{exaone-deep}, QwQ-32B / 32B-Preview \citep{qwq32b}, OpenThinker2-7B / 32B \citep{openthoughts}, and Light-R1-7B/14B/32B \citep{wen2025light}. To explore the reasoning capabilities of the frontier reasoning models, we also evaluated OpenAI's \texttt{O3-mini-medium} as well as DeepSeek's \texttt{DeepSeek-R1} models on TTT-Bench. Our evaluation framework is based on the DeepScaler \citep{deepscaler2025} and VeRL \citep{sheng2024hybridflow} codebases. All of our evaluations are done with AMD Instinct MI300X$^{TM}$ GPUs. In the results for all the benchmarks, we report Pass@1 scores that are calculated based on $k=16$ responses, where we generate $k$ responses for each test sample with a temperature of 0.6, a top-p value of 0.95, and a response length of 28k tokens, and calculate its corresponding Pass@1 as:
\begin{equation}
Pass@1\space = \frac{1}{k}\sum_{i=1}^{k}p_i
\end{equation}
Here, $p_i$ represents the correctness of the $i^{th}$ response ($p_i=1$ in the case of questions with multiple solutions if the model predicts either of the solutions correctly). The final solution predictions are extracted from the generated responses using regular expression extraction from the last boxed answer \citep{deepscaler2025}. Only for frontier models, we report the Pass@1 scores based on just one response per test sample in the case of TTT-Bench, whereas we report their corresponding published results in the case of math benchmarks.
\begin{table}[t]
\centering
\resizebox{\columnwidth}{!}{%
\begin{tabular}{l|llll|ll}
\hline
               & oTTT & dTTT & cTTT & sTTT & MATH 500 & AIME 2024 \\ \hline
o3-mini-medium & 79.6 & 83.0 & 52.2 & 75.0 & 97.3*     & 79.6*      \\
DeepSeek-R1    & 88.9 & 72.0 & 58.9 & 75.8 & 97.3*     & 79.8*      \\ \hline
\end{tabular}%
}
\caption{Frontier LRM Pass@1 performance on TTT-Bench and math benchmarks. The (*) indicates that the results for math benchmarks we sourced from the respective model's published results. The TTT-Bench task performances are based on a single response per test sample.}
\label{tab:frontier-performance}
\end{table}

\begin{figure*}[!ht]
\centering
  \includegraphics[width=\textwidth]{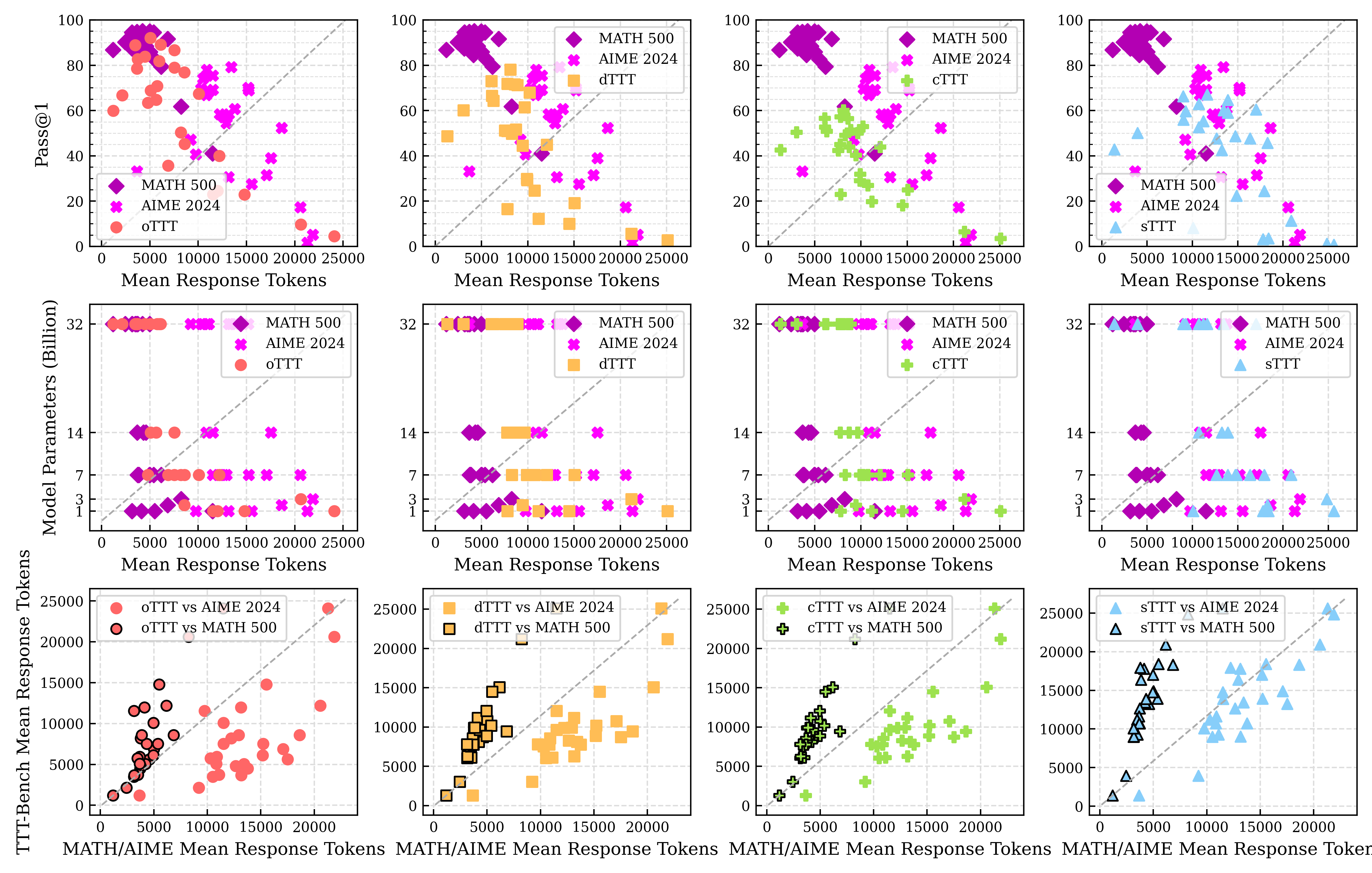}
  \caption{\textbf{Top:} Pass@1 performance of LRMs on MATH 500, TTT-Bench \& AIME 2024 against the mean length of the corresponding solution responses. \textbf{Mid:} Model size against the corresponding model's mean response lengths on MATH 500, TTT-Bench \& AIME 2024. \textbf{Bottom:} The mean response lengths used for solving the TTT-Bench task against the response lengths for solving math benchmarks for different models.}
  \label{fig:ctx-len}
\end{figure*}
\subsection{TTT-Bench Evaluation Results}
\paragraph{Weak reasoning ability of LRMs on simple and intuitive tasks:}
In Figure \ref{fig:test-reasoning-v-math}, we report the performance of SOTA reasoning models individually on the four games in TTT-Bench as well as on AIME 2024 \& MATH 500. In our analysis, we leverage the performance drop  ($\Delta  Pass@1$ as defined below) between the TTT-Bench tasks ($x$) and math benchmarks ($y$) as a metric to measure the divergence in reasoning ability of these models on different domain reasoning tasks:
\begin{equation}
\Delta Pass@1= Pass@1_{x}-Pass@1_{y}    
\end{equation}
Given the simplicity of the TTT-Bench tasks, we expect Pass@1 scores of SOTA LRMs to be higher relative to their performance on math benchmarks (positive $\Delta  Pass@1$), especially for models that do well on math benchmarks. \textbf{Surprisingly, we observed the opposite -- the majority of the evaluated models had lower Pass@1 scores on TTT-Bench compared to their Pass@1 scores on MATH 500 ($\downarrow$ Avg $\Delta  Pass@1$: -41.36\%) \& AIME 2024 ($\downarrow$ Avg $\Delta  Pass@1$: -4.88\%).} In the case of MATH 500, which consists of high school math problems, the drops are significant relative to all the TTT-Bench tasks across all the models, especially for the small models. Whereas in the case of the olympiad-level math benchmark AIME 2024, LRMs performed well on the oTTT task relative to AIME, with an $\uparrow$ Avg $\Delta  Pass@1$ of +8.56\%, whereas they struggled on other TTT-Bench tasks. The $\Delta  Pass@1$ trends also illustrate the relative task reasoning difficulties over the TTT-Bench tasks, which are consistent across $\Delta  Pass@1$ comparisons against both MATH 500 \& AIME 2024: oTTT<dTTT<sTTT<cTTT. We attribute this relative task difficulty order to the following two properties of these games: 1) Spatial constraints in the tasks, and 2) the number of possibilities to choose the next best move in the tasks. In Table \ref{tab:frontier-performance}, we also report the $Pass@1$ performance scores on TTT-Bench of flagship frontier models like \texttt{o3-mini-medium} \& \texttt{DeepSeek-R1}. Due to resource constraints, we only evaluated these models with one response per test sample of TTT-Bench and reported their respective published numbers for MATH 500 \& AIME 2024 \footnote{\href{o3-mini results}{https://openai.com/index/openai-o3-mini/}} \footnote{\href{DeepSeek-R1 results}{https://huggingface.co/deepseek-ai/DeepSeek-R1}}. From our results, we observe that even the frontier reasoning models find the TTT-Bench tasks as difficult as the standard math benchmarks. \textit{These performance trends indicate poor reasoning ability of SOTA LRMs on TTT-Bench tasks that are simple, intuitive, and of low complexity, compared to their performance on both MATH 500 \& AIME 2024 questions that are difficult and require relatively more knowledge and complex reasoning efforts.}

\paragraph{Reasoning models struggle with simple long-term strategic reasoning:}
Since TTT-bench is generated using an automated pipeline that leveraged a heuristic-based algorithm (Appendix \ref{sec:appendix-algo}) for generating the solutions and the respective game's final verdict ("Win", "Blocked", or "Fork"), we investigate the type of reasoning tasks these LRMs are good at by analyzing their performance over these these individual solution verdict category questions. As shown in Figure \ref{fig:test-category}, we observe that across the TTT-Bench tasks oTTT, dTTT, and sTTT, the performance of LRMs over questions that have the solution with verdict "Win" is consistently higher than the questions with verdict "Blocked", with the lowest for the questions with verdict "Fork". Whereas in the case of cTTT, the performance remained equally low for all the questions, independent of the solution verdict. This consistently high performance over questions with a solution verdict "Win" indicates that almost all the LRMs are capable of doing sound short-term thinking, where the solution is straightforward. On the contrary, a consistent dip in performance over slightly complex reasoning scenarios with "Blocked" \& "Fork" solution verdicts indicates that the LRMs struggle to successfully solve simple long-term strategic reasoning tasks where the models need to explore more possibilities and think strategically to either block the opponent's win or to identify a position which can lead to a long-term assured win.
\begin{figure*}[!ht]
\centering
  \includegraphics[width=\textwidth]{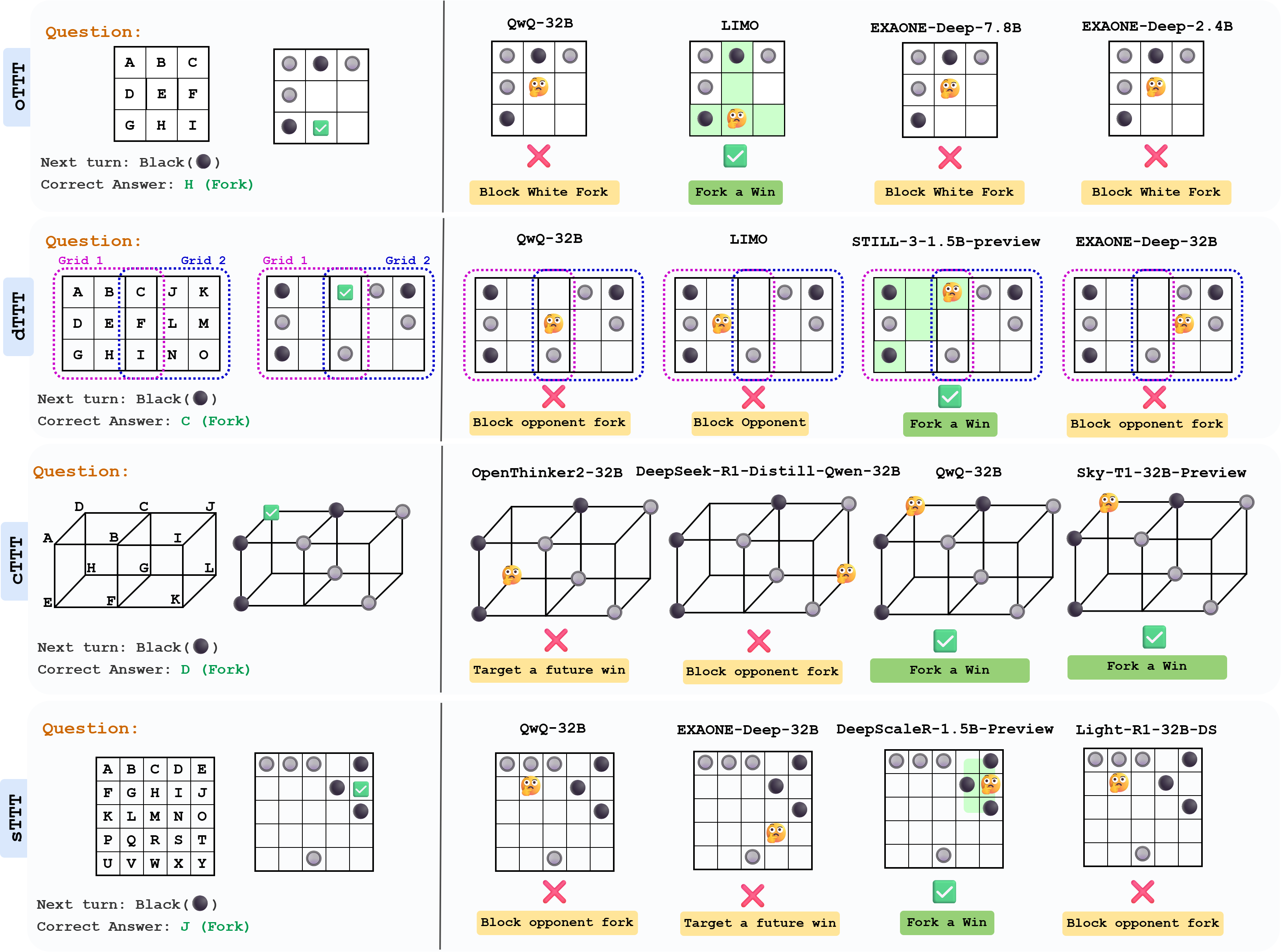}
  \caption{Examples of predicted solutions by SOTA LRMs for questions in TTT-Bench.}
  \label{fig:tttbench-example}
\end{figure*}
\paragraph{LRMs overthink on TTT-Bench, with an increase in model size resulting in improved performance and efficient use of chain-of-thought:}
Reasoning models are usually trained with long-context data to enable a long chain-of-thought (COT) that has proven to evoke emergent reasoning capabilities like self-reflection and backtracking \citep{deepseekai2025deepseekr1incentivizingreasoningcapability}. Ideally, a model should use a longer response length with more thinking tokens for questions deemed difficult and requiring a complex thinking process, whereas a shorter response length is used for simple and trivial questions \citep{jin-etal-2024-impact}. In Figure \ref{fig:ctx-len}, we compare 1) the task performance of the models against the mean length of the corresponding solution responses (\textbf{Top}), 2) the model scale against the corresponding mean response lengths (\textbf{Mid}), and 3) mean response lengths used for solving TTT-Bench task against the mean response lengths for solving math benchmarks for different models (\textbf{Bottom}). From these plots, we observe that LRMs solve MATH 500 questions with higher accuracy and shorter COT, whereas in the case of both AIME and TTT-Bench questions (especially dTTT, cTTT, \& sTTT), models used longer COT with high variability and lower performance. This is counterintuitive, as the question is that TTT-Bench tasks are relatively simple, straightforward, and don't require as much thinking as is required to solve AIME questions. Consequently, we find that the larger models achieve higher performance using shorter COT across all the benchmarks, where the order of response lengths used for solving different benchmarks is: MATH 500 < TTT-Bench < AIME 2024. A head-to-head comparison between the generated solution response lengths for solving TTT-Bench tasks $vs$ math benchmarks reveal that all the LRMs use longer COT for questions in TTT-Bench tasks against the MATH 500 questions, whereas they generate somewhat similar length COT when compared against AIME questions (especially in the case of sTTT against AIME). These findings suggest that these reasoning models consume similar amount thinking tokens for solving TTT-Bench tasks (even though they are trivial for humans) as used for solving olympiad-level math questions, and are found to produce long COT with circular, inconclusive, and repetitive thinking process for these trivial questions (examples of solution traces are provided in Appendix \ref{sec:reasoning-trace}), further indicating their inability to do simple, intuitive, and straightforward tasks. 

% \subsection{TTT-Bench Failure Mode Discussion}
To further understand the failure modes of SOTA LRMs, in Figure \ref{fig:tttbench-example} we illustrate a few question examples from TTT-Bench and the answers predicted by different models, with their main reasons for those predictions based on the reasoning traces. These examples shed light on the simplicity of our benchmarks as well as how these models reach incorrect or suboptimal TTT-Bench predictions. In the oTTT examples, majority of models predict position \texttt{"E"} as the next move, which although is a decent move to block the white ($\circ$) stones from forming a fork (\texttt{"A-E-I" \& "D-E-F"}), but they fail to identify a better move (\texttt{"H"}) that can guarantee a win for the current player by creating a fork for itself (\texttt{"B-E-H" \& "G-H-I"}) to ensure a win. Again, in the case of dTTT, cTTT \& sTTT, the models tend to predict a move that blocks a forking possibility of the opponent, rather than thinking towards a long-term win by placing a move that establishes a fork for itself and ensures a win. \textbf{These examples demonstrate that most of the time, SOTA LRMs produce suboptimal solutions on TTT-Bench, and struggle at long-term strategic planning towards the optimal outcome.}

\section{Conclusion}
In this work, we introduce \textbf{TTT-Bench}, a novel benchmark designed to assess the reasoning capabilities of LRMs on intuitive and low-complexity reasoning tasks, which are effortless for humans to solve from a young age. While these models achieve remarkable success on challenging mathematical benchmarks like MATH 500 \& AIME 2024, our evaluations reveal a consistent and surprising shortfall in their performance on simple two-player games, highlighting a fundamental gap in their reasoning capabilities. Through TTT-Bench, which is generated using our proposed automated and verifiable benchmark generation framework, we demonstrate that existing LRMs tend to overthink trivial tasks, exhibit performance degradation with longer outputs, and often miss optimal strategies, even when they possess the underlying capabilities to solve significantly harder problems. This discrepancy underscores the importance of diversifying evaluation protocols for reasoning models to include intuitive and simple benchmarks beyond the conventional STEM-focused paradigms.

\section{Limitations}
While \textbf{TTT-Bench} offers new insights into the reasoning capabilities of LRMs, several limitations must be acknowledged:

\paragraph{Narrow game domain.}
TTT-Bench focuses exclusively on a specific class of simple two-player board games inspired by Tic-Tac-Toe. Although these games are intuitive and require basic spatial and strategic reasoning, they represent only a small slice of the broader spectrum of intuitive reasoning tasks. As such, the conclusions drawn from TTT-Bench may not directly generalize to other forms of intuitive reasoning, such as commonsense planning, physical reasoning, or visual-spatial puzzles.

\paragraph{Text-only formulation.}
All TTT-Bench tasks are presented in purely textual form, even though the reasoning involved is highly spatial. While this setup allows for scalable and consistent evaluation using language models, it may introduce unnecessary difficulty due to the lack of visual context. Some model failures might stem from difficulty in mentally visualizing board states rather than limitations in reasoning per se.

\paragraph{Model access and decoding constraints.}
Some evaluations—especially for closed-source or API models—are limited to a single sample per query, unlike open models where multiple generations are averaged. This discrepancy may affect the comparability of results and lead to under- or over-estimation of certain models' performance.
\bibliography{custom}

\appendix
\section{TTT-Bench Examples}\label{Appendix-Examples}
Here we provide the questions from TT-Bench and the solution illustrations for the examples in Figure \ref{fig:tttbench}. For the solutions, we also provide the verdict and the states contributing to the corresponding verdict.
\begin{tcolorbox}[breakable,fonttitle=\bfseries\ttfamily,fontupper=\ttfamily,boxrule=1pt,width=\columnwidth,title={oTTT test sample example in Figure 1}]
        \textbf{Question}:

        Alice and Bob are playing a game on a 3x3 grid. The points on the grid are labeled top to bottom, left to right, as A,B,C,D,E,F,G,H,I. Alice plays white. Bob plays black. At each turn, the player places a stone of the corresponding color onto one of the positions that has not been occupied. Whoever has three stones in a line (horizontal, vertical, or diagonal) wins. Alice first places a white stone at A. Bob places a black stone at B. Alice places a white stone at H. Bob places a black stone at C. Where should Alice play next?

        \textbf{Game Visualization:}
        
        \includegraphics[width=0.6\columnwidth]{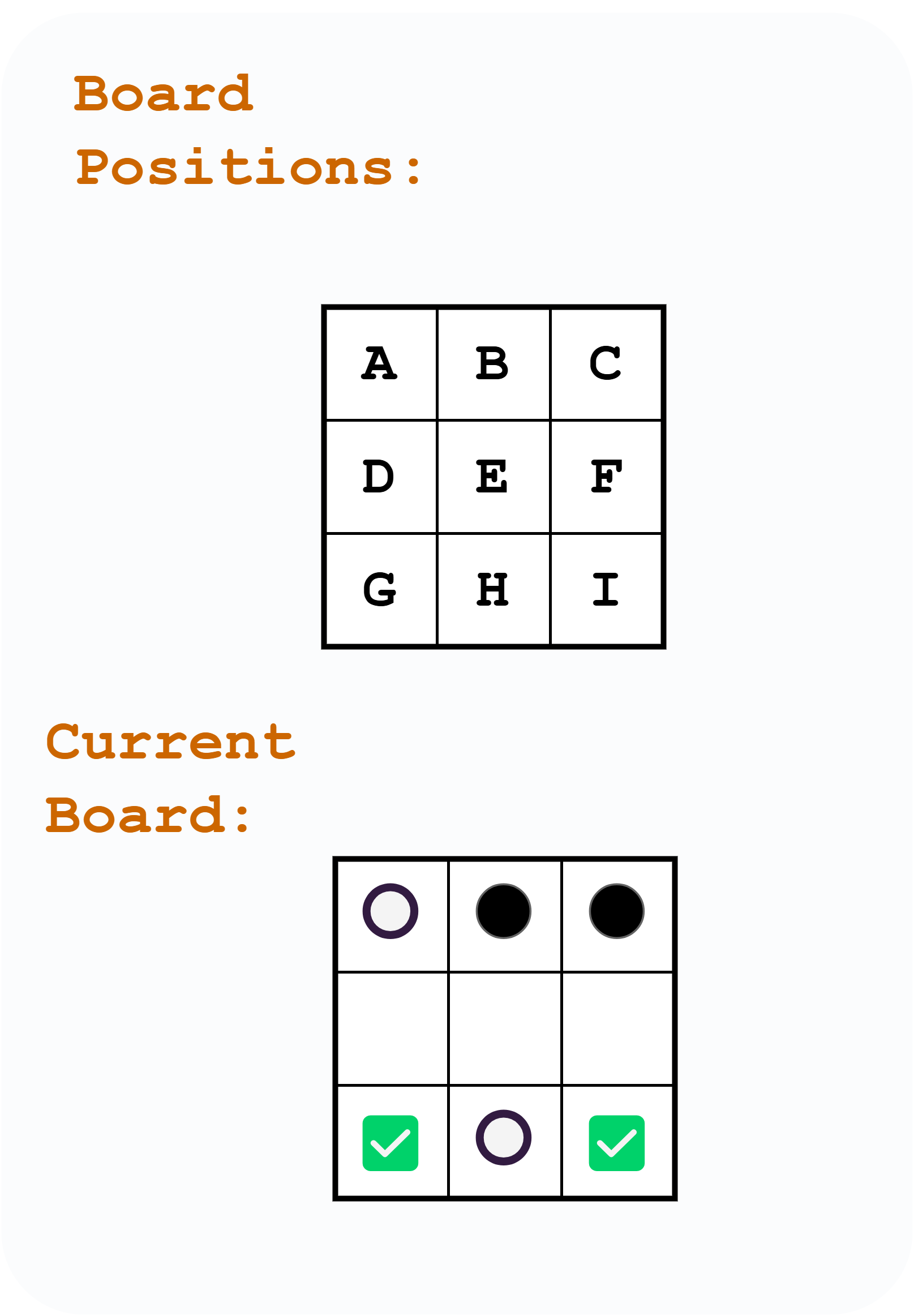}
        
        \textbf{Solution:}
        
        G (Fork with A-D-G \& G-H-I) or I (Fork with A-E-I \& G-H-I)
\end{tcolorbox}
\begin{tcolorbox}[breakable,fonttitle=\bfseries\ttfamily,fontupper=\ttfamily,boxrule=1pt,width=\columnwidth,title={dTTT test sample example in Figure 1}]
        \textbf{Question}:
        
        Alice and Bob are playing a game on two adjacent 3x3 grids. The points on the first grid are labeled top to bottom, left to right, as A,B,C,D,E,F,G,H,I. The points on the second grid are labeled top to bottom, left to right, as C,J,K,F,L,M,I,N,O. Note that points C,F, I are shared by the two grids. Alice plays white. Bob plays black. At each turn, the player places a stone of the corresponding color onto one of the positions that has not been occupied. Whoever has three stones in a line (horizontal, vertical, or diagonal) on either grid wins. Alice first places a white stone at I. Bob places a black stone at J. Then Alice at F. Bob places a black stone at L. Then Alice at B. Bob places a black stone at H. Where should Alice play next?

        \textbf{Game Visualization:}
        
        \includegraphics[width=0.6\columnwidth]{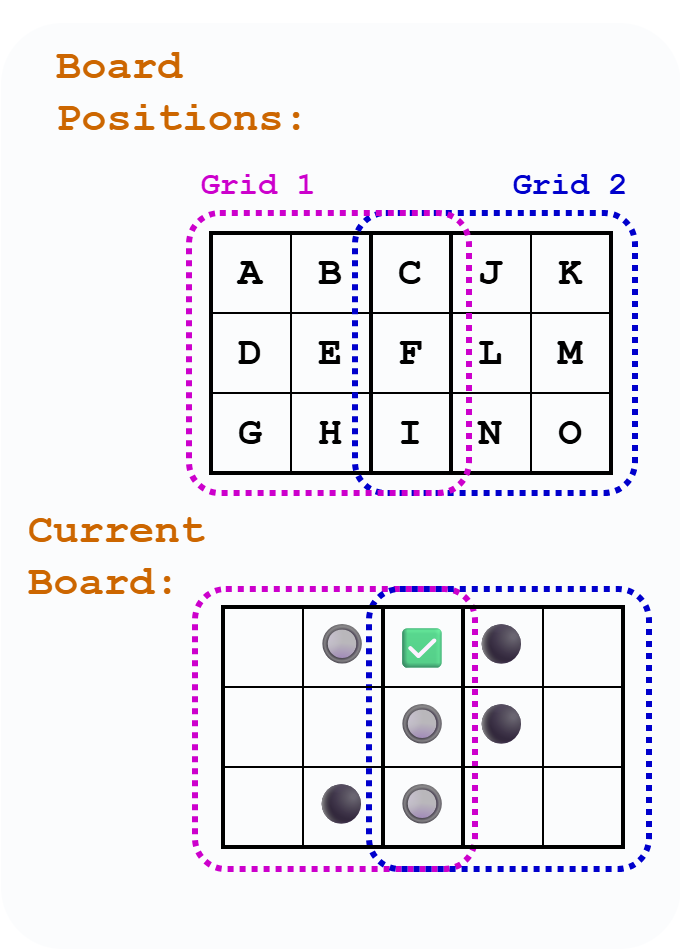}
        
        \textbf{Solution:}
        
        C (Win with C-F-I)
\end{tcolorbox}
\begin{tcolorbox}[breakable,fonttitle=\bfseries\ttfamily,fontupper=\ttfamily,boxrule=1pt,width=\columnwidth,title={cTTT test sample example in Figure 1}]
        \textbf{Question}:
        
        Alice and Bob are playing a game on two adjacent cubes. ABCD forms the top rectangle in the first cube, and BIJC forms the top rectangle in the second cube. EFGH forms the bottom rectangle in the first cube, and FKLG forms the bottom rectangle in the second cube. AE is an edge, BF is an edge, CG is an edge, DH is an edge, IK is an edge, and JL is an edge. Note that vertices B,C,G,F are shared by the two cubes. Alice and Bob play a game where they take turns to put stickers on the vertices of the cubes that have not been occupied. Alice plays white stickers. Bob plays black stickers. The person who has four stickers on the same plane on either cube wins. Alice first places a white stone at A. Then Bob places a black stone at B. Then Alice places a white stone at C. Then Bob places a black stone at I. Then Alice places a white stone at E. Then Bob places a black stone at K. Where should Alice play next?

        \textbf{Game Visualization:}
        
        \includegraphics[width=0.6\columnwidth]{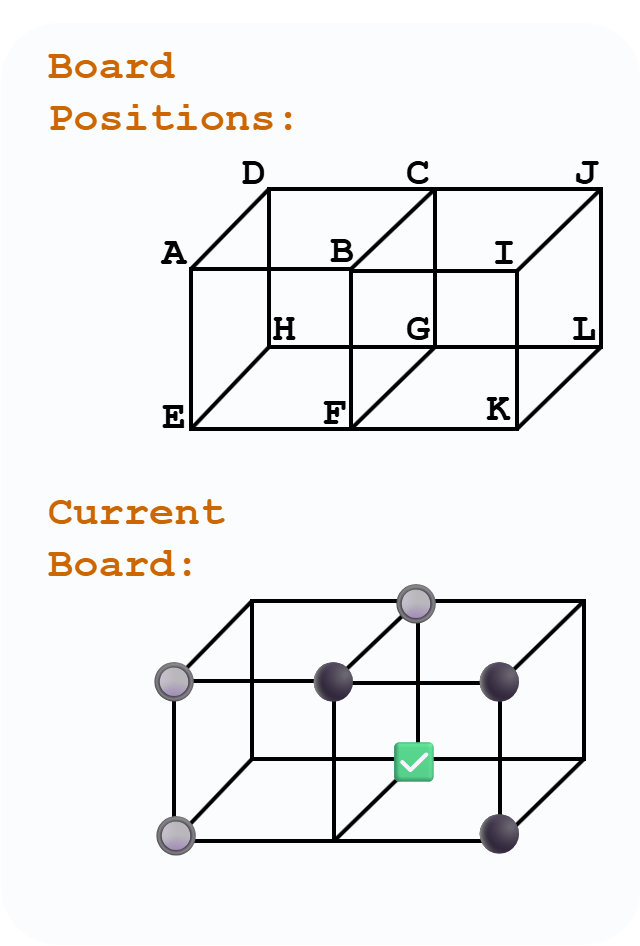}
        
        \textbf{Solution:}
        
        G (Win with A-C-G-E)
\end{tcolorbox}
\begin{tcolorbox}[breakable,fonttitle=\bfseries\ttfamily,fontupper=\ttfamily,boxrule=1pt,width=\columnwidth,title={sTTT test sample example in Figure 1}]
        \textbf{Question}:
        
        Alice and Bob are playing a game on a board. There are 5 equally spaced horizontal lines where the distance between two neighboring horizontal lines is 1. Similarly, there are 5 equally spaced vertical lines, and the distance between two neighboring vertical lines is 1. There are 25 intersection points between the 5 horizontal lines and 5 vertical lines. These 25 points are labeled from top to bottom, left to right, as A, B, C, D, …, Y. Alice plays white. Bob plays black. At each turn, the player places a stone of the corresponding color onto one of the 25 points that have not been occupied. Whoever has four stones that form either a unit square (with side length of 1) or a “diagonal square” with side length equal to the square root of 2 wins. For example, ABGF is a unit square. FBHL is a diagonal square. Alice first places a white stone at A. Then Bob places a black stone at D. Then Alice places a white stone at B. Then Bob places a black stone at K. Then Alice places a white stone at H. Then Bob places a black stone at U. Where should Alice play next?

        \textbf{Game Visualization:}
        
        \includegraphics[width=0.6\columnwidth]{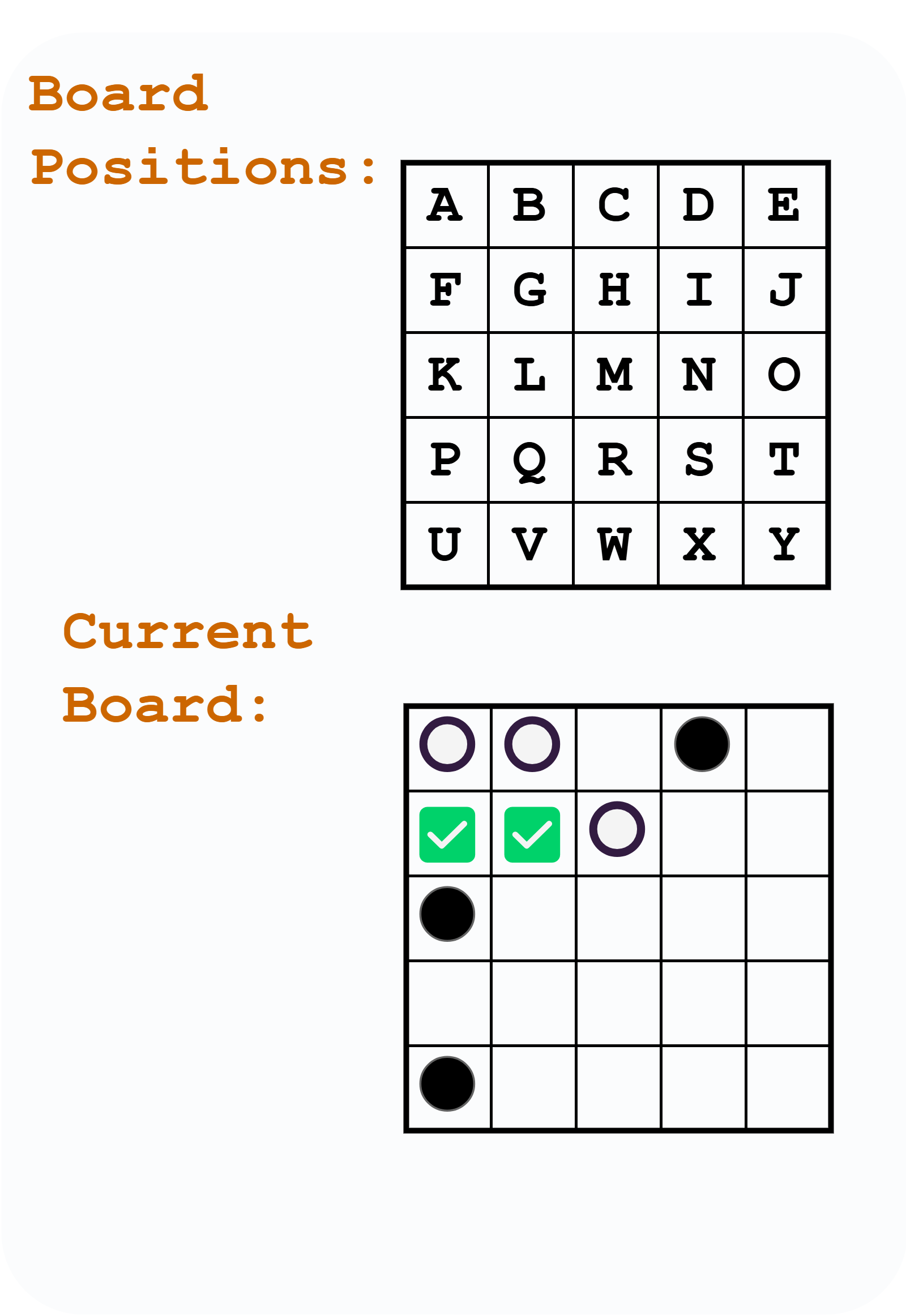}
        
        \textbf{Solution:}
        
        F (Fork with A-B-F-G \& F-B-H-L) or G (Fork with A-B-G-F \& A-C-H-G)
\end{tcolorbox}
\section{Solution Trace Examples} \label{sec:reasoning-trace}
Here we provide the reasoning traces produced by different models for question examples in Figure \ref{fig:tttbench}.
\begin{tcolorbox}[enhanced jigsaw,breakable,colback=green!5!white,colframe=green!75!black, fonttitle=\bfseries\ttfamily,fontupper=\ttfamily,boxrule=1pt,width=\columnwidth,title={LIMO Solution Trace for oTTT question in Figure 6 (Correct)}]
        \textbf{Question:}
        
        Alice and Bob are playing a game on a 3x3 grid. The points on the grid are labeled top to bottom, left to right, as A,B,C,D,E,F,G,H,I. Alice plays white. Bob plays black. At each turn, the player places a stone of the corresponding color onto one of the positions that has not been occupied. Whoever has three stones in a line (horizontal, vertical, or diagonal) wins. Alice first places a white stone at A. Bob places a black stone at B. Alice places a white stone at C. Bob places a black stone at G. Alice places a white stone at D. Where should Bob play next? Let's think step by step and output the final answer within \textbackslash boxed\{\}.

        \textbf{Correct Answer:} H (Forms fork with B-E-H \& G-H-I)

        \textbf{Game Visualization:}
        
        \includegraphics[width=0.8\columnwidth]{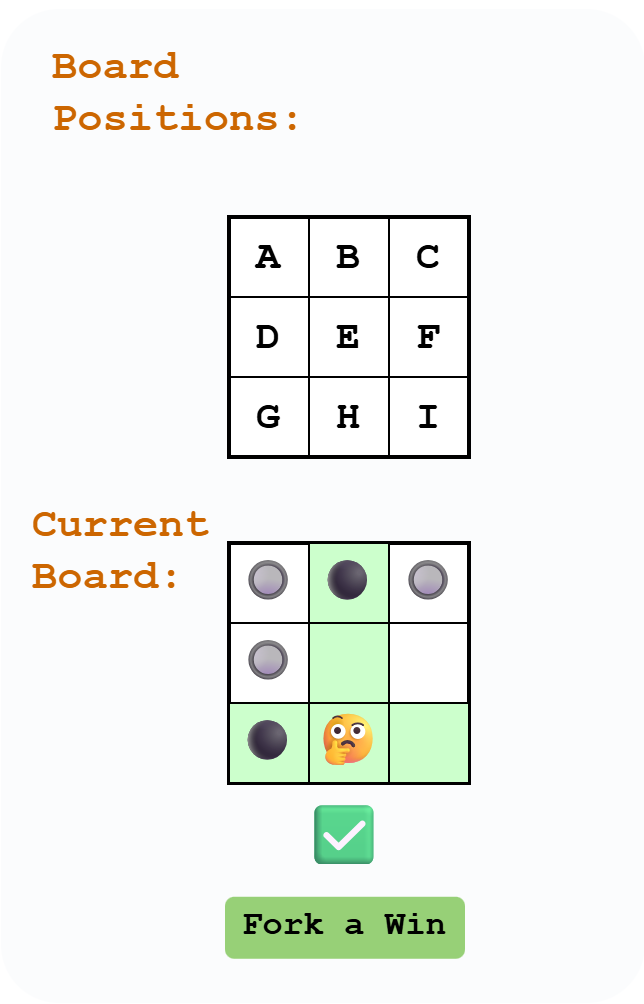}
        
        \textbf{Solution:}
        
        <think>
        
        ...
        
        </think>
        
        Therefore, if Bob plays at H, Alice must play at I to prevent Bob from winning. Then, Bob plays at F, leading to a draw. If Alice makes a mistake and plays at F, Bob wins. Therefore, playing at H is a good move because it puts pressure on Alice to play correctly. If she doesn't, Bob wins. If she does, the game is a draw.

        Therefore, considering all options, Bob should play at H. This creates a threat that Alice must respond to, and if she responds incorrectly, Bob can win. If she responds correctly, the game is a draw, which is still a better outcome than losing.

        Therefore, the best move for Bob is to play at H.

        **Final Answer**
        
        \textbackslash boxed\{H\}
        
\end{tcolorbox}

\begin{tcolorbox}[enhanced jigsaw,breakable,colback=red!5!white,colframe=red!75!black, fonttitle=\bfseries\ttfamily,fontupper=\ttfamily,boxrule=1pt,width=\columnwidth,title={QwQ-32B Solution Trace for dTTT question in Figure 6 (Wrong)}]
        \textbf{Question:}
        
        Alice and Bob are playing a game on two adjacent 3x3 grid. The points on the first grid are labeled top to bottom, left to right, as A,B,C,D,E,F,G,H,I. The points on the second grid are labeled top to bottom, left to right, as C,J,K,F,L,M,I,N,O. Note that points C,F, I are shared by the two grids. Alice plays white. Bob plays black. At each turn, the player places a stone of the corresponding color onto one of the positions that has not been occupied. Whoever has three stones in a line (horizontal, vertical, or diagonal) on either grid wins. Alice first places a white stone at I. Bob places a black stone at G. Then Alice at J. Bob places a black stone at K. Then Alice at D. Bob places a black stone at A. Then Alice at M. Where should Bob play next? Let's think step by step and output the final answer within \textbackslash boxed\{\}.
        
        \textbf{Correct Answer:} C (Forms fork with A-B-C \& G-E-C)

        \textbf{Game Visualization:}
        
        \includegraphics[width=0.8\columnwidth]{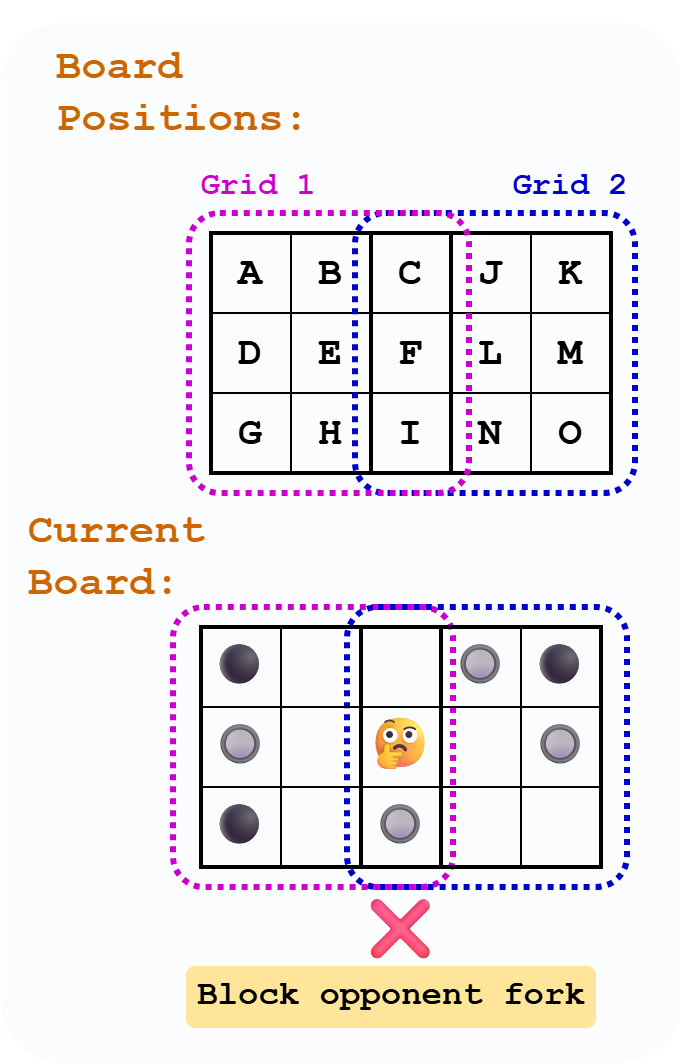}
        
        \textbf{Solution:}
        
        <think>
        
        ...
        
        </think>
        
        Alice and Bob are playing a game on two adjacent 3x3 grids. The key steps and reasoning are as follows:
        
        1. **Current Positions**:
        
           - **Grid1**:
           
             - Alice: I (bottom-right), D (middle-left)
             
             - Bob: G (bottom-left), A (top-left)
             
           - **Grid2**:
           
             - Alice: J (top-middle), M (middle-right), I (bottom-left)
             
             - Bob: K (top-right)

        2. **Alice's Potential Threats**:
        
           - In **Grid1**, Alice has D and I. To form a line, she needs a third stone in a row, column, or diagonal.
           
           - In **Grid2**, Alice has J, M, and I. These positions do not form a line yet, but she could aim for a diagonal or column.

        3. **Critical Blocking Moves**:
        
           - **Grid1's Column3 (C, F, I)**: Alice has I. If she takes C and F, she can form a vertical line. Bob can block this by taking F.
           
           - **Grid1's Row2 (D, E, F)**: Alice has D. Taking F would block her from completing this row.
           
           - **Grid2's Diagonal (K-L-I)**: If Alice takes L, she could form a diagonal with I and L. However, Bob's K blocks this.

        4. **Strategic Considerations**:
        
           - Taking F in **Grid1** blocks both the vertical column3 and the horizontal row2 threats.
           
           - Taking F also affects **Grid2's column1**, preventing Alice from forming a vertical line there.

        After evaluating all possibilities, the optimal move for Bob is to place a stone at **F**, which blocks Alice's potential threats in both grids effectively.
        
        \textbackslash boxed\{F\}
        
\end{tcolorbox}

\begin{tcolorbox}[enhanced jigsaw, breakable,colback=red!5!white,colframe=red!75!black,fonttitle=\bfseries\ttfamily,fontupper=\ttfamily,boxrule=1pt,width=\columnwidth,title={DeepSeek-R1-Distill-Qwen-32B Solution Trace for cTTT question in Figure 6 (Wrong)}]
        \textbf{Question:}
        
        Alice and Bob are playing a game on two adjacent cubes. ABCD forms the top rectangle in the first cube and BIJC forms the top rectangle in the second cube. EFGH forms the bottom rectangle in the first cube and FKLG forms the bottom rectangle in the second cube. AE is an edge, BF is an edge, CG is an edge, DH is an edge, IK is an edge, and JL is an edge. Note that vertices B,C,G,F are shared by the two cubes. Alice and Bob plays a game where they take turns to put stickers on the vertices of the cubes that have not been occupied. Alice plays white stickers. Bob plays black stickers. The person who has four stickers on the same plane on either cube wins. Alice first places a white stone at B. Then Bob places a black stone at A. Then Alice places a white stone at G. Then Bob places a black stone at C. Then Alice places a white stone at J. Then Bob places a black stone at E. Then Alice places a white stone at K. Where should Bob play next? Let's think step by step and output the final answer within \textbackslash boxed\{\}.
        
        \textbf{Correct Answer:} D (Forms fork with A-D-H-E \& D-C-E-F)

        \textbf{Game Visualization:}
        
        \includegraphics[width=0.8\columnwidth]{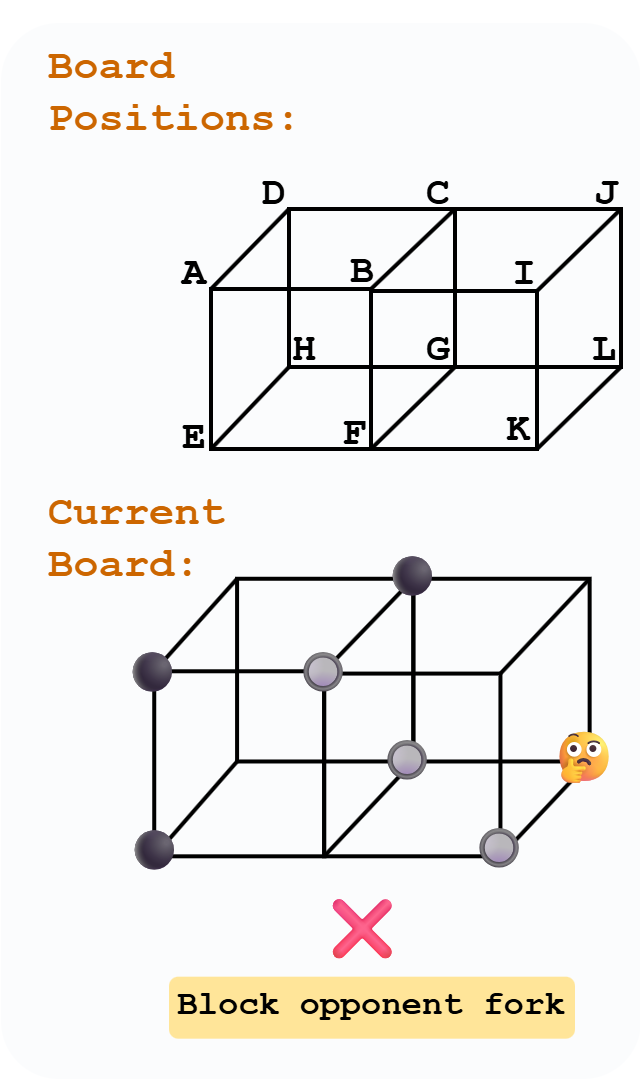}
        
        \textbf{Solution:}
        
        <think>
        
        ...
        
        </think>
        
        Alice and Bob are playing a game on two adjacent cubes. The goal is to place stickers on vertices such that four stickers of the same color form a plane on either cube. Alice goes first and places white stickers, while Bob places black stickers. The current moves are as follows:

        1. Alice places a white sticker at B.
        
        2. Bob places a black sticker at A.
        
        3. Alice places a white sticker at G.
        
        4. Bob places a black sticker at C.
        
        5. Alice places a white sticker at J.
        
        6. Bob places a black sticker at E.
        
        7. Alice places a white sticker at K.

        The vertices involved are A, B, C, D, E, F, G, H, I, J, K, L. The shared vertices are B, C, F, G. Alice's current stickers are at B, G, J, K. Bob's current stickers are at A, C, E. The remaining vertices are D, H, I, L, F.

        To determine where Bob should place his next sticker, we need to consider the potential planes where Alice could win and block her. Alice is threatening multiple planes, but Bob can only block one.

        After analyzing the potential planes and considering the current state of the game, Bob should place his next sticker at L to block Alice's potential win on the back face (C, G, L, J) and the bottom face (F, K, L, G). This move also blocks Alice 
        +from completing these planes, preventing her from winning on her next turn.

        Thus, the final answer is:
        
        \textbackslash boxed\{L\}
        
\end{tcolorbox}

\begin{tcolorbox}[enhanced jigsaw,breakable,colback=red!5!white,colframe=red!75!black, fonttitle=\bfseries\ttfamily,fontupper=\ttfamily,boxrule=1pt,width=\columnwidth,title={QwQ-32B Solution Trace for sTTT question in Figure 6 (Wrong)}]
        \textbf{Question:}
        
        Alice and Bob are playing a game on a board. There are 5 equally spaced horizontal lines where the distance between two neighboring horizontal lines is 1. Similarly, there are 5 equally spaced vertical lines, and the distance between two neighboring vertical lines is 1. There are 25 intersection points between the 5 horizontal lines and 5 vertical lines. These 25 points are labeled from top to bottom, left to right, as A, B, C, D, …, Y.  Alice plays white. Bob plays black. At each turn, the player places a stone of the corresponding color onto one of the 25 points that have not been occupied. Whoever has four stones that form either a unit square (with side length of 1)  or a “diagonal square” with side length equal to the square root of 2 wins. For example, ABGF is a unit square. FBHL is a diagonal square. Alice first places a white stone at A. Then Bob places a black stone at E. Then Alice places a white stone at B. Then Bob places a black stone at I. Then Alice places a white stone at C. Then Bob places a black stone at O. Then Alice places a white stone at W. Where should Bob play next? Let's think step by step and output the final answer within \textbackslash boxed\{\}.
        
        \textbf{Correct Answer:} J (Forms fork with D-E-J-I \& I-J-O-N)

        \textbf{Game Visualization:}
        
        \includegraphics[width=0.8\columnwidth]{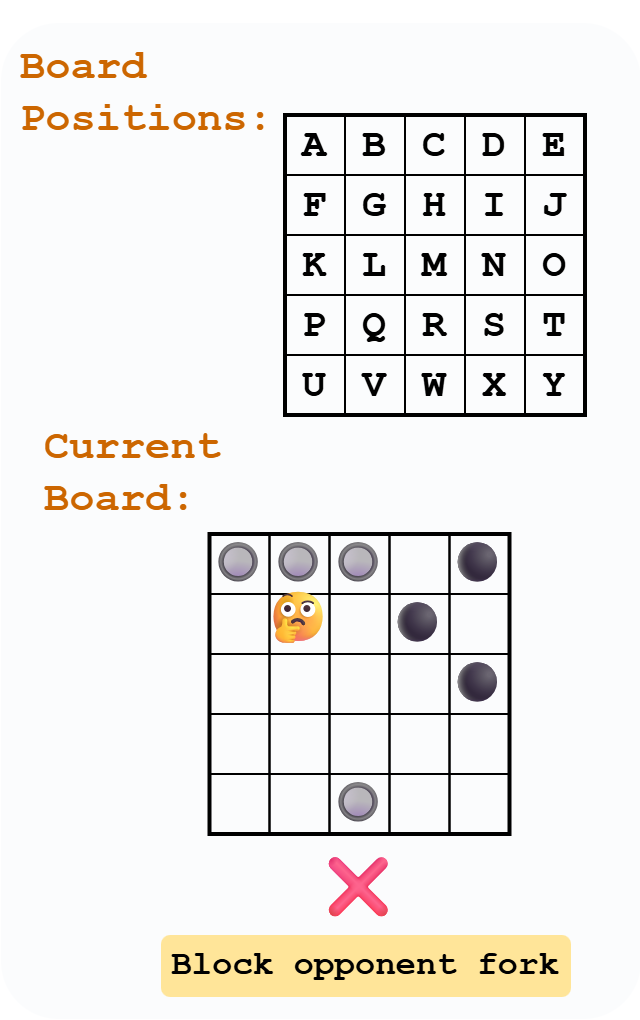}
        
        \textbf{Solution:}
        
        <think>
        
        ...
        
        </think>
        
        Alice and Bob are playing a game on a 5x5 grid where the goal is to form a unit square (side length 1) or a diagonal square (side length square root 2) with four stones. Alice's moves so far are at A, B, C, and W, while Bob's moves are at E, I, and O.

        Key considerations:
        
        1. **Alice's stones**: A(0,0), B(0,1), C(0,2), W(4,2).
        
        2. **Bob's stones**: E(0,4), I(1,3), O(2,4).
        
        3. **Threat analysis**: Alice's top-row stones (A, B, C) could form part of a unit square if combined with stones below. The critical points for such squares include G(1,1) and H(1,2).
        
        4. **Blocking potential squares**: The unit square involving A, B, and needing F(1,0) and G(1,1) is a key threat. Since Alice has A and B, blocking G(1,1) would prevent her from completing this square in future moves.

        After evaluating all potential squares and critical points, the most strategic move for Bob to block Alice's potential threats is to place at G(1,1).
        
        \textbackslash boxed\{G\}
        
\end{tcolorbox}
\section{Algorithm for generating TTT-Bench solutions}\label{sec:appendix-algo}
On the next page, we provide the detailed algorithm used for the automated generation of TTT-Bench questions and their corresponding solutions.
\section{Performance Results} \label{appendix-results}
In Table \ref{main-result-table} and Table \ref{catergory-results} we report the Pass@1 performance numbers used in Figure \ref{fig:test-reasoning-v-math} and Figure \ref{fig:test-category} respectively.

\clearpage
\begin{algorithm*}
\small
    \caption{Algorithm for generating TTT-Bench solutions}\label{aglo:ttt-bench}
    \DontPrintSemicolon
    \textbf{Total number of positions:} $T$,
    \textbf{Number of moves played:} $N=n_c+n_o$,
    \textbf{Number of positions in winning lines:} $n$,
    \textbf{Number of winning lines:} $t$,
    \textbf{Current player positions:} $P^c:\{p_1^c,...,p_{n_c}^c\}$,
    \textbf{Opponent positions:} $P^o:\{p_1^o,...,p_{n_o}^o\}$,
    \textbf{Available Position:} $A:\{a_1,...,a_{T-N}\}$,
    \textbf{Wining States:} $W:\{(w_1^1,...,x_n^1),...(w_1^t,...,x_n^t)\}$

    \SetKwFunction{Win}{check\_won}
    \SetKwProg{Fn}{Function}{:}{}
    \Fn{\Win{$P_{new}^c$}}{
        \For{$(w_1^i,...,x_n^i)$ in $W$}{
            \If{$\forall( \space w_j^i\in P_{new}^c | \space w_j^i \in (w_1^i,...,x_n^i))$}{
                  return True;
               }
        }
        return False;
    }
    \SetKwFunction{CheckWin}{get\_wins}
    \SetKwProg{Fn}{Function}{:}{}
    \Fn{\CheckWin{$P^c,A$}}{
        $w\_moves \leftarrow \{\}$ \\
        \For{$a_i$ in $A$}{
            $P_{new}^c \leftarrow P^c + a_i$ \\
            \If{\Win{$P_{new}^c$}}{
                $w\_moves \leftarrow w\_moves+a_i$
            }
        }
        return $w\_moves$;
    }
    \SetKwFunction{CheckFork}{get\_forks}
    \SetKwProg{Fn}{Function}{:}{}
    \Fn{\CheckFork{$P^c,A$}}{
        $f\_moves \leftarrow \{\}$ \\
        \For{$a_i$ in $A$}{
            $P_{new}^c \leftarrow P^c + a_i \space , \space A^{potential} \leftarrow a_j \in A; a_j\neq a_i \space , \space threat\_count \leftarrow 0$ \\
            \For{$(w_1^i,...,x_n^i)$ in $W$}{
                $W^{current} \leftarrow w_j^i \space | \space w_j^i \in (w_1^i,...,x_n^i) \space \And \space w_j^i \in P_{new}^c$ \\
                $W^{opponent} \leftarrow w_j^i \space | \space w_j^i \in (w_1^i,...,x_n^i) \space \And \space w_j^i \notin P_{new}^c \space \And \space w_j^i \notin A^{potential}$ \\
                \If{$len(W^{current})==n-1 \space \And \space len(W^{opponent})==0$}{
                    $threat\_count \leftarrow threat\_count+1$ \\
                }
            }
            \If{$threat\_count>=2$}{
                $f\_moves \leftarrow f\_moves+a_i$
            }
        
        }
        return $f\_moves$;
    }
    \SetKwFunction{Solution}{get\_solution}
    \SetKwProg{Fn}{Function}{:}{}
    \Fn{\Solution{$P^c,P^o,A$}}{
        $w\_moves \leftarrow $\CheckWin{$P^c$,$A$} \Comment{\textcolor{magenta}{Check if current player has wining moves.}} \\
        \If{$len(w\_moves)>=1$}{
            return $w\_moves$,"Win";
        }
        $opp\_w\_moves \leftarrow $\CheckWin{$P^o$,$A$} \Comment{\textcolor{magenta}{Check if current player has blocking moves.}} \\
        \If{$len(opp\_w\_moves)==1$}{
            return $opp\_w\_moves$,"Blocked";
        }\Else{return None;}
        $f\_moves \leftarrow $\CheckFork{$P^c$,$A$} \Comment{\textcolor{magenta}{Check if current player has forking moves.}}\\
        \If{$len(f\_moves)>=1$}{
            return $f\_moves$,"Fork";
        }
    }
\end{algorithm*}

\begin{table*}[!ht]
    \centering
    \resizebox{\textwidth}{!}{
    \begin{tabular}{r|c|c|c|c|c|c}
    \toprule
        \textbf{Models} & \textbf{oTTT} & \textbf{dTTT} & \textbf{cTTT} & \textbf{sTTT} & \textbf{AIME 2024} & \textbf{MATH 500} \\ \hline
        QwQ-32B & 92.03 & 78.13 & 60.14 & 60.10 & 79.17 & 95.00 \\ 
        QwQ-32B-Preview & 66.73 & 60.13 & 50.49 & 50.16 & 47.08 & 90.16 \\ 
        EXAONE-Deep-32B & 89.09 & 71.25 & 51.46 & 60.42 & 70.00 & 94.95 \\ 
        OpenThinker2-32B & 88.85 & 73.00 & 56.60 & 66.30 & 74.17 & 94.53 \\ 
        DeepSeek-R1-Distill-Qwen-32B & 70.77 & 51.19 & 42.36 & 55.31 & 69.58 & 93.28 \\ 
        s1.1-32B & 83.76 & 71.81 & 57.22 & 62.86 & 60.63 & 92.46 \\ 
        Light-R1-32B-DS & 81.80 & 71.56 & 56.46 & 67.03 & 77.92 & 94.69 \\ 
        LIMO & 78.55 & 64.31 & 50.83 & 55.89 & 58.33 & 91.91 \\ 
        OpenThinker-32B & 82.66 & 66.56 & 52.71 & 59.69 & 68.33 & 93.83 \\ 
        Sky-T1-32B-Preview & 59.87 & 48.69 & 42.57 & 42.86 & 33.13 & 86.75 \\ \hline
        DeepSeek-R1-Distill-Qwen-14B & 68.81 & 50.94 & 45.00 & 52.66 & 66.88 & 93.45 \\ 
        Light-R1-14B-DS & 78.92 & 61.50 & 50.14 & 59.06 & 75.42 & 94.04 \\ 
        s1.1-14B & 64.77 & 51.69 & 43.89 & 42.50 & 38.96 & 88.38 \\ \hline 
        OpenThinker2-7B & 63.48 & 49.75 & 49.10 & 47.66 & 57.08 & 92.16 \\ 
        EXAONE-Deep-7.8B & 86.64 & 67.94 & 52.99 & 64.64 & 68.96 & 94.54 \\ 
        AceMath-RL-Nemotron-7B & 67.40 & 44.94 & 43.89 & 48.65 & 69.17 & 93.76 \\ 
        DeepSeek-R1-Distill-Qwen-7B & 45.22 & 29.25 & 31.88 & 47.66 & 54.38 & 92.05 \\ 
        OpenThinker-7B & 35.66 & 24.63 & 27.01 & 22.40 & 31.46 & 85.90 \\ 
        Light-R1-7B-DS & 50.31 & 29.81 & 29.03 & 24.48 & 58.33 & 92.18 \\ 
        s1.1-7B & 39.95 & 19.19 & 25.00 & 11.35 & 17.29 & 79.45 \\ \hline 
        EXAONE-Deep-2.4B & 76.84 & 44.50 & 40.35 & 45.68 & 52.29 & 91.60 \\ 
        s1.1-3B & 9.68 & 5.63 & 6.53 & 1.20 & 5.21 & 61.81 \\ \hline
        DeepSeek-R1-Distill-Qwen-1.5B & 22.92 & 10.06 & 18.19 & 3.49 & 27.50 & 82.58 \\ 
        DeepScaleR-1.5B-Preview & 23.04 & 16.50 & 22.99 & 8.18 & 40.63 & 87.43 \\ 
        s1.1-1.5B & 4.47 & 2.69 & 3.47 & 0.78 & 1.67 & 41.10 \\ 
        STILL-3-1.5B-preview & 24.51 & 12.25 & 19.79 & 3.18 & 30.63 & 84.59 \\
    \bottomrule
    \end{tabular}}
    \caption{Pass@1 performance of LRMs on TTT-Bench, AIME 2024, and MATH500.}
    \label{main-result-table}
\end{table*}

\begin{table*}[!ht]
    \centering
    \resizebox{\textwidth}{!}{
    \begin{tabular}{r|c|c|c|c|c|c|c|c|c|c|c|c}
    \toprule
        & \multicolumn{3}{c|}{\textbf{oTTT}} & \multicolumn{3}{c|}{\textbf{dTTT}} & \multicolumn{3}{c|}{\textbf{cTTT}} & \multicolumn{3}{c}{\textbf{sTTT}} \\ \cline{2-13} 
        \textbf{Models} & \textbf{Win} & \textbf{Blocked} & \textbf{Fork} & \textbf{Win} & \textbf{Blocked} & \textbf{Fork} & \textbf{Win} & \textbf{Blocked} & \textbf{Fork} & \textbf{Win} & \textbf{Blocked} & \textbf{Fork} \\ \hline
        QwQ-32B & 98.12 & 95.77 & 82.39 & 96.09 & 95.62 & 51.41 & 63.12 & 72.29 & 49.53 & 75.62 & 91.25 & 36.77 \\ 
        QwQ-32B-Preview & 85.31 & 53.12 & 53.69 & 83.12 & 49.69 & 42.34 & 44.69 & 51.25 & 52.81 & 77.08 & 53.33 & 35.1 \\ 
        EXAONE-Deep-32B & 99.84 & 94.3 & 67.61 & 95.62 & 90.31 & 37.34 & 42.19 & 66.46 & 44.84 & 96.46 & 66.46 & 39.38 \\ 
        OpenThinker2-32B & 99.38 & 95.77 & 66.19 & 95.47 & 95.31 & 39.38 & 49.69 & 68.75 & 50.94 & 92.29 & 86.25 & 43.33 \\ 
        DeepSeek-R1-Distill-Qwen-32B & 87.5 & 67.46 & 51.99 & 70.94 & 55.31 & 29.38 & 43.75 & 45.0 & 39.69 & 87.92 & 69.79 & 31.77 \\ 
        s1.1-32B & 97.03 & 82.72 & 65.62 & 94.38 & 83.12 & 43.59 & 52.19 & 63.96 & 54.69 & 86.88 & 78.12 & 43.23 \\ 
        Light-R1-32B-DS & 95.78 & 81.43 & 61.93 & 95.0 & 85.62 & 41.09 & 55.62 & 66.25 & 49.53 & 88.75 & 89.38 & 45.0 \\ 
        LIMO & 93.75 & 75.0 & 61.36 & 86.25 & 63.75 & 42.66 & 36.56 & 59.17 & 51.72 & 82.71 & 71.46 & 34.69 \\ 
        OpenThinker-32B & 95.47 & 87.5 & 58.81 & 92.34 & 82.81 & 32.66 & 42.5 & 66.25 & 47.66 & 90.21 & 75.42 & 36.56 \\ 
        Sky-T1-32B-Preview & 76.88 & 46.32 & 51.42 & 65.47 & 51.25 & 30.63 & 33.12 & 44.79 & 45.62 & 62.71 & 44.79 & 31.98 \\ \hline
        DeepSeek-R1-Distill-Qwen-14B & 85.0 & 65.26 & 49.72 & 75.31 & 52.19 & 25.94 & 29.69 & 51.46 & 47.81 & 82.5 & 62.92 & 32.6 \\ 
        Light-R1-14B-DS & 94.06 & 78.86 & 59.38 & 91.56 & 70.62 & 26.88 & 42.19 & 56.46 & 49.38 & 91.25 & 73.96 & 35.52 \\ 
        s1.1-14B & 83.44 & 60.48 & 42.33 & 76.88 & 46.56 & 29.06 & 26.25 & 49.58 & 48.44 & 65.0 & 51.04 & 26.98 \\ \hline
        OpenThinker2-7B & 83.28 & 57.17 & 41.48 & 73.75 & 35.0 & 33.12 & 31.87 & 54.17 & 53.91 & 77.29 & 47.92 & 32.71 \\ 
        EXAONE-Deep-7.8B & 96.09 & 94.85 & 64.2 & 90.0 & 77.19 & 41.25 & 38.75 & 57.92 & 56.41 & 95.0 & 86.25 & 38.65 \\ 
        AceMath-RL-Nemotron-7B & 87.5 & 59.93 & 47.16 & 71.09 & 36.25 & 23.12 & 44.06 & 38.12 & 48.12 & 82.92 & 65.0 & 23.33 \\ 
        DeepSeek-R1-Distill-Qwen-7B & 57.66 & 34.56 & 40.91 & 44.38 & 19.38 & 19.06 & 27.81 & 30.0 & 35.31 & 57.92 & 28.96 & 14.17 \\ 
        OpenThinker-7B & 43.44 & 31.62 & 30.4 & 35.47 & 15.31 & 18.44 & 17.81 & 21.25 & 35.94 & 31.46 & 21.04 & 18.54 \\ 
        Light-R1-7B-DS & 71.72 & 36.76 & 35.51 & 47.34 & 21.25 & 16.56 & 24.06 & 27.92 & 32.34 & 52.71 & 24.17 & 10.52 \\ 
        s1.1-7B & 50.16 & 36.58 & 29.83 & 25.16 & 15.0 & 15.31 & 17.19 & 21.04 & 31.87 & 15.83 & 9.58 & 10.0 \\ \hline
        EXAONE-Deep-2.4B & 77.97 & 97.79 & 53.41 & 57.5 & 56.25 & 25.62 & 24.06 & 53.33 & 38.75 & 54.58 & 86.67 & 20.73 \\ 
        s1.1-3B & 9.53 & 7.72 & 13.07 & 5.62 & 3.12 & 6.88 & 5.31 & 2.92 & 9.84 & 1.46 & 0.62 & 1.35 \\ \hline 
        DeepSeek-R1-Distill-Qwen-1.5B & 28.59 & 18.01 & 21.31 & 10.0 & 6.88 & 11.72 & 15.0 & 14.37 & 22.66 & 5.83 & 3.96 & 2.08 \\ 
        DeepScaleR-1.5B-Preview & 40.62 & 8.46 & 15.62 & 19.69 & 11.56 & 15.78 & 22.19 & 17.08 & 27.81 & 14.17 & 6.04 & 6.25 \\ 
        s1.1-1.5B & 3.44 & 4.41 & 6.25 & 3.28 & 0.62 & 3.12 & 12.81 & 13.12 & 28.28 & 1.04 & 1.46 & 0.31 \\
        STILL-3-1.5B-preview & 31.09 & 16.54 & 24.43 & 15.00 & 7.19 & 12.03 & 12.81 & 13.12 & 28.28 & 6.04 & 1.88 & 2.40 \\ 
    \bottomrule
    \end{tabular}}
    \caption{Pass@1 performance of LRMs over individual solution verdict category ("Win", "Blocked", \& "Fork") questions in TTT-Bench tasks.}
    \label{catergory-results}
\end{table*}
\end{document}